\documentclass{article}

\usepackage[preprint]{neurips_2024}

\usepackage[utf8]{inputenc} 
\usepackage[T1]{fontenc}    
\usepackage{hyperref}       
\usepackage{url}            
\usepackage{booktabs}       
\usepackage{amsfonts}       
\usepackage{nicefrac}       
\usepackage{microtype}      
\usepackage{xcolor}     

\usepackage{xcolor}         
\usepackage{amsmath}
\usepackage{algorithm,algorithmic}
\usepackage{mathtools}

\usepackage{graphicx} 
\usepackage{caption}
\usepackage{subcaption}
\usepackage{hyperref}
\usepackage{url}
\usepackage{natbib}
\usepackage{wrapfig}
\usepackage{mathtools}

\newcommand{\regions}{\mathsf{M}}

\newcommand{\duration}{\mathsf{T}}
\newcommand{\kernel}{\mathsf{K}}
\newcommand{\conditions}{\mathsf{C}}

\newcommand{\btt}[1]{\bm{\mathtt{#1}}}

\usepackage{amsmath,amsfonts,bm}

\def\eqref#1{equation~\ref{#1}}

\def\1{\bm{1}}

\def\eps{{\epsilon}}

\def\vc{{\bm{c}}}

\def\vh{{\bm{h}}}

\def\vq{{\bm{q}}}
\def\vr{{\bm{r}}}

\def\vx{{\bm{x}}}
\def\vy{{\bm{y}}}

\DeclareMathAlphabet{\mathsfit}{\encodingdefault}{\sfdefault}{m}{sl}
\SetMathAlphabet{\mathsfit}{bold}{\encodingdefault}{\sfdefault}{bx}{n}
\newcommand{\tens}[1]{\bm{\mathsfit{#1}}}
\def\tA{{\tens{A}}}

\def\tH{{\tens{H}}}

\def\tX{{\tens{X}}}
\def\tY{{\tens{Y}}}

\newcommand{\E}{\mathbb{E}}

\newcommand{\R}{\mathbb{R}}

\definecolor{blue1}{RGB}{0,128,255}
\definecolor{blue3}{RGB}{0,0,128}
\definecolor{darkpastelgreen}{rgb}{0.01, 0.75, 0.24}
\definecolor{cerulean}{rgb}{0.0, 0.48, 0.65}

\title{Robust and highly scalable estimation of directional couplings from time-shifted signals}

\author{%
  Louis Rouillard$~^*$ \\
  Parietal, Inria Saclay \\
  Université Paris-Sud \\
  \texttt{louis.rouillard@gmail.com} \\
  \And
  Luca Ambrogioni \thanks{Equal contribution} \\
  Donders Institute for Brain, Cognition, and Behaviour\\
  Radboud University\\
  \texttt{luca.ambrogioni@donders.ru.nl} \\
  \AND
  Demian Wassermann \\
  Parietal, Inria Saclay \\
  Université Paris-Sud \\
  \texttt{demian.wassermann@inria.fr} \\
}

\begin{document}
\maketitle
\begin{abstract}
The estimation of directed couplings between the nodes of a network from indirect measurements is a central methodological challenge in scientific fields such as neuroscience, systems biology and economics. Unfortunately, the problem is generally ill-posed due to the possible presence of unknown delays in the measurements. In this paper, we offer a solution of this problem by using a variational Bayes framework, where the uncertainty over the delays is marginalized in order to obtain conservative coupling estimates. To overcome the well-known overconfidence of classical variational methods, we use a hybrid-VI scheme where the (possibly flat or multimodal) posterior over the measurement parameters is estimated using a forward KL loss while the (nearly convex) conditional posterior over the couplings is estimated using the highly scalable gradient-based VI. In our ground-truth experiments, we show that the network provides reliable and conservative estimates of the couplings, greatly outperforming similar methods such as regression DCM.
\end{abstract}

\section{Introduction}
Several physical, biological and technological complex systems can be characterized by the structure of interconnections between a large number of their relatively simple components. For example, the human brain is composed of approximately 100 billion neurons, which are connected by potentially as many as 600 trillion synapses \cite{herculano2009human, von2016search, loomba2022connectomic}. Similarly, the working of an individual cell is depends on a complex networks of bio-chemical interactions that can again be understood in terms of causal interactions between its components \cite{white2005signaling, kravchenko2012fundamental}. The scientific study of such complex systems fundamentally relies on the use of non-invasive high-coverage measuring methods. Unfortunately, these methods tend to perform indirect measurements that often induces variable time delays in the recorded signals. For example, fMRI measures neuronal activity through changes in concentration of oxygenated blood, which introduces a temporal shift due to the latency of the neural-metabolic coupling \cite{buckner1998event, lindquist2009modeling}. These potentially variable time delays pose a serious problem when the aim is to estimate directed couplings, since the potential time reversal of the cross-correlograms can lead to spurious inference \cite{stephan2012short, ramsey2010six, raut2019time}. 

In this paper, we introduce an inference method that uses a mixture of forward and reversed variational inference losses to recover the complex multimodal joint posterior of coupling and shift variables. We use a forward amortized variational inference (FAVI) approach to estimate the posterior distributions of the shift parameters \citep{papamakarios2016fast, ambrogioni2019forward}; while we use a highly scalable gradient-based reverse KL approach to estimate the (nearly convex) posterior of the coupling matrix given the shifts \citep{mnih2016variational, kucukelbir2017automatic, frassle2017regression}. Treating the shift parameters differently from the coupling parameters allows us to avoid the 'catastrophic' collapse of uncertainty due to the mode seeking behavior of the reversed KL, while keeping a very high level of scalability. By marginalizing out the uncertainty over the shift variables, we obtain a well-calibrated posterior distribution over the couplings. The approach can be applied on networks with hundreds of nodes and minimizes the risk of spurious inference due to differential shifts, thereby solving one of the fundamental problems that affected effective connectivity methods in neuroscience and other fields. 

\section{Related Work}
In classical statistical signal processing, directed linear coupling are estimated using cross-correlograms \citep{barlow1959autocorrelation}, vector autoregressive modeling \citep{zivot2006vector}, Granger causality \citep{granger1969investigating, shojaie2022granger} and non-parametric spectral decomposition methods \citep{west2020measuring}. All these methods crucially depend on time-shifts in the cross-correlation between signals, which are assumed to be caused by delayed causal couplings. The problem of estimating directed couplings from time-shifted signals gained substantial attention in neuroscience due to the fact that large scale measurements of neural signal are often indirect, relying on complex processes such as neural-hemodynamic coupling in fMRI \citep{buckner1998event, lindquist2009modeling} or calcium-influx in calcium imaging \citep{grienberger2012imaging}. Since the measured signals depend on complex, and possibly variable, biological processes, a direct use of cross-correlogram or autoregressive-based methods could lead to spurious results \citep{ramsey2010six}. Because temporal delays were considered unreliable, directed couplings were initially estimated from individual temporally-averaged measurements by varying an external stimulus (or condition) and observing the resulting changes in the correlation structure between different nodes. This form of analysis used \emph{structural equation modeling} methods (SEM) and is particularly appropriate when the temporal resolution of the signal is very low such as in PET scans \citep{laird2008modeling}. The temporal aspect of the coupling analysis was exploited with the introduction of \emph{dynamic causal models}, which assume an underlying deterministic dynamical system and a complex time-shifted emission model \citep{friston2003dynamic}. In this family of Bayesian techniques, the posterior distribution is estimated using variational Bayes approach with Laplace approximation of the likelihood \citep{friston2003dynamic}. The approach was extended to the study of resting state dynamics by including stochastic inputs to the nodes \citep{daunizeau2012stochastic}. Unfortunately, these models could only be scaled to small networks of selected nodes. Recently, regression DCM achieved high scalability by using linear emission models and gradient-based inference in the frequency domain \citep{frassle2017regression}. The current work is a development on the multivariate dynamical systems (MDS) approach \citep{ryali2011multivariate}. MDS models assume that the latent dynamic is driven by a state equation. This latent state generates (potentially delayed) observations through an 'observation equation'. MDS has been extensively tested in neural simulations \citep{ryali2016multivariate} and optogenetic 
experiments \citep{ryali2016combining}. The use of forward KL in directed coupling analysis was introduced in \citep{ambrogioni2019spikecake} for spike measurements. However, n this work the measurements did not introduce a delay.

\section{Methods}
\label{sec:methods}
Our goal is to infer the directed statistical coupling between latent signals. As an example, the signals can represent the activity of activity in neural populations in the brain coupled through axonal pathways. We assume that the measurements are generated by the convolution of the latent signals with response functions with variable temporal shifts and spectral properties. This is often the case in neural measurements, where neural activities is often inferred from time-shifted proxy-signals such as the haemodynamic (BOLD) response in fMRI measurements of brain activity \citep{lindquist2009modeling}. For the sake of simplicity and robustness, we assume the latent coupling between regions to be linear and at the latent activation level. The variability in the time-shifts shifts in the response functions can potentially change the temporal ordering of the cross-correlations because measurements with lower delay will predate measurements with higher delay. This can lead to spurious directed coupling estimates in cross-correlogram based methods, because a $A \rightarrow B$ coupling can be interpreted as $A \rightarrow B$ if the measurement of $A$ is more delayed compared to the measurement of $B$. 

In the following, we will outline a Bayesian model designed to integrate out the uncertainty introduced by the shifts, and thereby to reduce the risk of spurious inferences. 


\begin{figure}
    \centering
    \includegraphics[width=\textwidth]{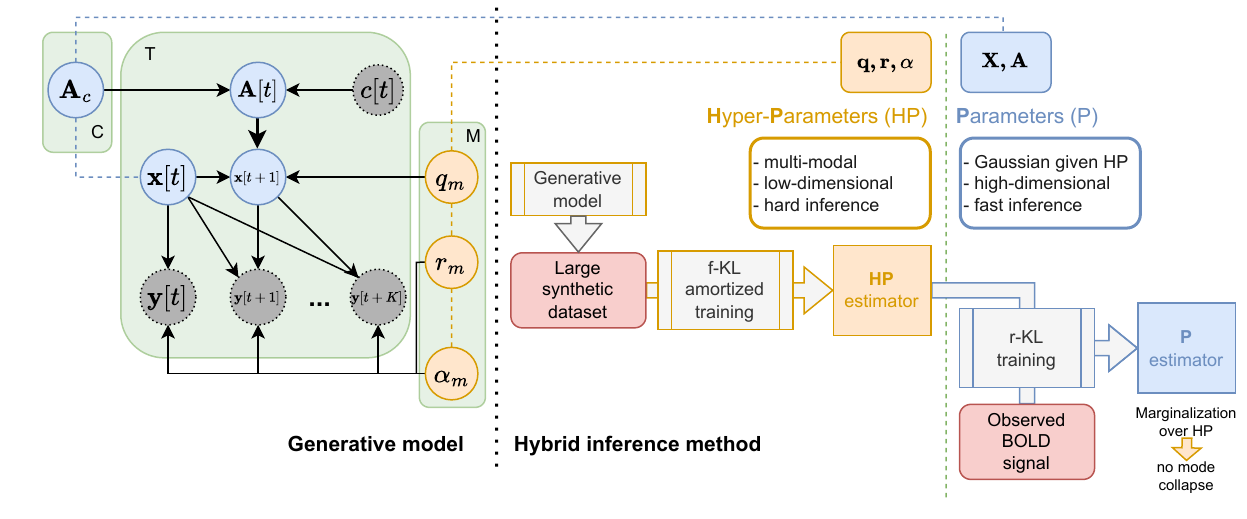}
    \caption{
        Graphical representation for the MDS model, and general principle of our hybrid method.
        Parameters are separated into two groups.
        For the hyper-parameters (HP), we use the forward KL to obtain a well-calibrated estimator.
        The HP estimator is plugged into a scalable reverse-KL training to estimate the parameters (P).
    }
    \label{fig:MDS-graphical}
\end{figure}

\paragraph{Latent activation dynamics}
Latent activations $\tX$ are subject to the directed coupling between different regions. We assume this coupling to be linear and parameterized by a coupling matrix $\tA$. The evolution of the latent signal follows the linear Gaussian state-space model \citep{mdsi_method}:
\begin{equation}
    \label{eq:x}
    \begin{aligned}
        \vx[t+1] &= \tA[t] ~\vx[t] + \bm{\eps} \\
        \bm{\eps} &\sim \mathcal{N}(\mathbf{0}, \vq) \\
        \vq &\in \R^{+ \regions}
    \end{aligned}
\end{equation}
where $\bm{\eps}$ denotes some latent white Gaussian noise that we assume to be independent and of different amplitude across regions.

\paragraph{Time-shifted measurements}
$\vy_m \in \R^\duration$ denotes the measurable time series for the location $m$, where $\duration$ denotes the temporal duration of the signal. We denote as $\regions$ the number of regions.
We model $\vy_m$ as the convolution of some latent activation by a response function (RF).
$\vx_m \in \R^\duration$ denotes the latent activation.
The RF is assumed to be location-specific: the location $m$ is associated with the RF $\vh_m \in \R^\kernel$ of temporal duration $\kernel$.
We obtain $\vy_m$ as:
\begin{equation}
    \label{eq:y}
    \begin{aligned}
        y_m[t] &= (\vh_m \ast \vx_m)[t] + \eta \\
        \eta &\sim \mathcal{N}(0, r_m) \\
        \vr &= \left[ r_1 \ \hdots \ r_\regions \right] \in \R^{+ \regions}
    \end{aligned}
\end{equation}
where $[t]$ denotes the time indexing, and $\bm{\eta} \sim \mathcal{N}(\mathbf{0}, \vr)$ denotes measurement white Gaussian noise that we assume to be independent and of different amplitude across regions. Following \cite{glover},  we model the location-specific RF $\vh_m$ as a linear combination of a base RF $\mathbf{h}_0$ and its time derivative $\mathbf{h}'_0$, both of temporal duration $\kernel$:
\begin{equation}
    \begin{aligned}
        \vh_m &= \cos (\alpha_m) \mathbf{h}_0 + \sin (\alpha_m) \dot{\mathbf{h}'_0} \\
        \alpha_m &\in \left] - \pi / 4, \pi / 4 \right[ \\
    \end{aligned}
\end{equation}
where following \citet{hrf_circle}, the RF coefficients are parameterized on the unit circle. This means that the RF for a location $m$ is entirely described by the angle $\alpha_m$. We model the parameters of the RF as independent across locations.
Considering all the locations at once, we respectively denote $\tY, \tX \in \R^{\regions \times \duration}$ and $\tH \in \R^{\regions \times \kernel}$ the concatenated observable signals, latent signals and RFs. Vectorizing the convolution operation across regions, we can write $\vy[t] = (\tH \ast \tX)[t] + \bm{\eta}$. A graphical representation of the MDS model is visible in \ref{fig:MDS-graphical}.

\subsection{Problem statement: inferring region coupling from the convolved signals}
\label{sec:inference}

\paragraph{Inferring parameters susceptible to yield the observed signal}
Given the MDS model described in Section~\ref{sec:methods}, the observed signal $\tY$ and experimental conditions $\vc$, we aim to infer the parameters susceptible to generating $\tY$.
The MDS model is associated with the joint distribution $p$, which factorizes as:
\begin{equation}
    \begin{aligned}
        p(\tY, \vc, \tX, \tA, \vq, \vr, \tH) &= p(\tY | \tX, \vr, \tH) \\
        &\ \times p(\tX|\vc, \tA, \vq) \\
        &\ \times p(\vc)p(\tA)p(\vr)p(\vq)p(\tH)
    \end{aligned}
\end{equation}
where $p(\vc)$ is a uniform categorical prior, $p(\tA)$ is a sparsity-inducing Laplace prior, $p(\tH)$ corresponds to a uniform prior over the angle $\alpha$ between the bounds $\left] - \pi / 4, \pi / 4 \right[$, and $p(\vq)$ and $p(\vr)$ are log-normal priors.
$p(\tX|\vc, \tA, \vq)$ and $p(\tY | \tX, \vr, \tH)$ correspond to the Normal distributions described in Eq. \ref{eq:x}\&\ref{eq:y}. Following the Bayesian inference formalism, we search for the posterior distribution of the coupling matrix: $p(\tA | \tY, \vc)$.
$p(\tA | \tY, \vc)$ denotes a distribution because there are several sources of uncertainty in the problem, and therefore $\tA$ cannot be inferred unequivocally.
In particular, both the latent and the observable noise levels are unknown. More importantly, the RF $\tH$ for the different regions is also unknown, which could induce a time reversal of the cross-correlograms between the observable signals at different locations. When estimating the latent signal $\tX$ and the coupling matrix $\tA$, we want to ensure that the uncertainty in all the other parameters is properly marginalized.
That is to say, we do not want to underestimate the uncertainty when inferring the parameters of interest. In detail, our method focuses on the proper marginalization of the RF $\tH$.
Each combination of different RFs for the locations yields ---via de-convolution--- a different set of latent signals $\tX$.
In turn, each different set of latent signals yields a different estimate for the coupling matrix $\tA$. Theoretically, the Bayesian framework allows weighting all those scenarios by their likelihood of generating the observed BOLD signal $\tY$.
This results in a single posterior distribution $p(\tA | \tY, \vc)$ that integrates all the sources of uncertainty in the problem. However, in practice, inference methods may fail to recover the true posterior $p(\tA | \tY, \vc)$, resulting in uncertainty underestimation and biased estimation.

\paragraph{A practical hurdle for inference: multiple modes explaining the observed data}
To explain how inference may fail in practice, we briefly overview its underlying mechanisms. Inference methods explore the high-dimensional parameter space$(\vq, \vr, \tH, \tX, \tA)$. In this large parameter space, several regions may explain well the observed signal $\tY$. For instance, different combinations of RFs and underlying signals.
Low-probability regions may separate those highly explanatory regions, creating separate distribution \textit{modes}.
Approximate inference method such as reversed gradient-based VI are prone to become stuck in one of those modes and ignore equally relevant sets of solutions to the problem. Multi-modality is indeed a known issue for off-the-shelf inference methods. In the context of Markov chain Monte Carlo (MCMC) methods, this can result in the non-mixing of multiple chains \citep{review_mcmc}. In the context of Variational Inference (VI), the phenomenon is known as mode collapse \citep{blei_review}. Critically, while inference methods may fail to recover the true uncertainty in $p(\tA | \tY, \vc)$, they still output the distribution corresponding to the mode they are stuck into.
This can be a misleading result: recovering a probabilistic output, experimenters may assume that \textit{all} the uncertainty in the problem has been captured.
Yet, in practice, off-the-shelf methods may only recover \textit{part} of the problem's uncertainty.
In the context of the MDS generative model --described in Section \ref{sec:methods}--- mode collapse can result in over-inflated statistical confidence when inferring the connections between locations, and even in spurious connections discovery.
In this paper, we propose a robust inference method to marginalize the uncertainty in the RF $\tH$ and the noise levels $\vq$ and $\vr$ properly when inferring the latent signal $\tX$ and the coupling matrix $\tA$.


\subsection{Hybrid Variational Bayes}
\label{sec:h-VB}

In this section, we describe our hybrid variational Bayes method (h-VB) to tackle the multi-modality in inference.
The term \textit{hybrid} refers to separating the parameters $(\vq, \vr, \tH, \tX, \tA)$ into two groups treated using different inference methods as illustrated in figure \ref{fig:MDS-graphical}. Specifically, as described below, we use a reverse-KL gradient-based VI loss for the coupling and latent signal parameters as they correspond to a well-behaved unimodal conditional optimization problem. On the other hand, we use a forward-amortized loss (FAVI) \cite{ambrogioni2019forward} for the noise and HR parameters since their posterior distribution is often highly multi-modal and the FAVI approach is capable of learning multi-modalities in the posterior. This results in a hybrid approach that combines the efficiency and scalability of reverse-KL VI for large-scale inference of large coupling matrices with the robustness of FAVI on a smaller set of key (hyper-)parameters. 

\paragraph{Inference using parameter optimization: the automatic differentiation variational inference (ADVI) framework}
Variational Bayes, also referred to as variational inference (VI), frames approximate inference as an optimization problem.
Inference reduces to choosing a variational family $\mathcal{Q}$ and finding inside that family the distribution $q(\vq, \vr, \tH, \tX, \tA; \phi) \in \mathcal{Q}$ closest to the unknown posterior $p(\vq, \vr, \tH, \tX, \tA | \tY, \vc)$.
To find the closest distribution $q$, we optimize the parameters $\phi$ to minimize a loss $\mathcal{L}$ that will be detailed in the next paragraphs.
To minimize $\mathcal{L}$, we proceed via gradient descent, which entitles being able to differentiate through $\mathcal{L}$.
Yet the loss $\mathcal{L}$ features expectation over parametric distributions, such as $q$.
This ill-posed differentiation prevents off-the-shelf optimization.
To circumvent this issue, we use the reparameterization trick \citep{kingma2013auto}: parametric distributions are reformulated as parametric transformations of fixed distributions ---such as the standard Normal distribution $\mathcal{N}(0, 1)$.
This reparameterization lets us differentiate through $\mathcal{L}$.
In turn, differentiating through $\mathcal{L}$ lets us leverage the powerful automatic differentiation libraries and optimizers developed in the deep learning community.
As a result, we can infer the parameters $(\vq, \vr, \tH, \tX, \tA)$ susceptible to producing the observed BOLD signal $\tY$ in a fast and scalable manner.

We optimize the variational distribution $q$ to approximate the unknown posterior $p(\vq, \vr, \tH, \tX, \tA | \tY, \vc)$.
We factorize $q$ into two densities:
\begin{equation}
    q(\vq, \vr, \tH, \tX, \tA; \phi) = q_\text{HP}(\vq, \vr, \tH; \phi_\text{HP}) \times q_\text{P}(\tX, \tA | \vq, \vr, \tH; \phi_\text{P})
\end{equation}
where $q_\text{HP}$ denotes our \textit{hyper-parameter} estimator, and $q_\text{P}$ our \textit{parameter} estimator.
Per our "hybrid" method, both factors are trained using different losses, as explained in the next two sections.

\paragraph{Hyper-parameter (HP) estimation}
Our main goal when training $q_\text{HP}$ is to avoid mode collapse, the phenomenon described in the \textit{practical hurdle} paragraph of \ref{sec:inference}.
Avoiding mode collapse will be ensured by the loss used for the training.
We consider the different regions as independent inference problems and factorize $q_\text{HP}$ as:
\begin{equation}
    q_\text{HP}(\vq, \vr, \tH | \tY = \btt{Y}; \phi_\text{HP}) = \prod_{m=1 .. \regions} q_\text{region}(q_m, r_m, \alpha_m ; f(\btt{y}_m; \phi_\text{HP}))
\end{equation}
where $q_\text{region}$ approximates a location's noise levels and RF given a realization of the region's abservable signal $\btt{y}$.
We use a Masked Autoregressive Flow \citep[MAF,][]{maf} to build $q_\text{region}$.
The flow approximates the joint distribution of $(\alpha, q, r)$.
To condition this distribution by the value of $\vy$, we feed to the flow an encoding of the observed region's observable signal $f(\btt{y}_m)$. As encoder $f$, we use a time convolutional neural network whose weights are jointly trained with the flow weights.
Combining a flow and a neural network encoder yields a very expressive density approximator able to model multimodal, heavy-tailed, and highly correlated distributions.

We train $q_\text{region}$ to minimize the forward Kullback-Leibler (f-KL) loss \citep{nf_review}, that is to say to maximize the probability of $(q, r, \alpha)$ given $\vy$:
\begin{equation}
    \begin{aligned}
        \phi_\text{HP}^*    &= \min_{\phi_\text{HP}} \mathcal{L}_\text{HP}^\text{f-KL} = \min_{\phi_\text{HP}} \E_{q, r, \alpha, \vy \sim p} \left[ - \log q_\text{region}(q, r, \alpha ; f(\vy ; \phi_\text{HP})) \right]
    \end{aligned}
\end{equation}
where the expectation $\E_{q, r, \alpha, \vy \sim p}$ denotes the training over a large synthetic dataset sampled from the generative model. The algorithm iterates between the following steps: \textbf{1)} it uses the generative model described in Section \ref{sec:methods} to sample the latent parameters $(\vq, \vr, \tH, \tX, \tA, \vc)$ and associated synthetic BOLD signal $\tY$; \textbf{2)} for each synthetic sample, it separates the BOLD signals $\btt{y}_m$ and parameters $(\alpha_m, q_m, r_m)$ corresponding to the different regions $m$; \textbf{3)} for each region $m$, it feeds $\btt{y}_m$ to the encoder $f$, and use the obtained encoding to condition our normalizing flow; \textbf{4)} it evaluates the probability of the parameters $(\alpha_m, q_m, r_m)$ under the conditioned flow; \textbf{5)} through gradient descent, it maximizes that probability, updating the weights of both the flow and the encoder.

After several epochs ---cycling through synthetic samples from the generative model--- $q_\text{region}$ converges to a good approximation of $p(q, r, \alpha | \vy)$. The training of $q_\text{region}$ is \textit{amortized}, which means that once trained, $q_\text{region}$ can estimate the hyper-parameters of \textit{any} brain region by feeding the region's BOLD signal $\btt{y}$ to the encoder $f$. We can then reuse $q_\text{region}$ across symmetrical inference problems inside the MDSI generative model, a concept named \textit{plate amortization} in the automatic VI literature \citep{pavi}.

\paragraph{Parameter (P) estimation}
Our main goal when training $q_\text{P}$ is inference speed and scalability.
This is due to the large dimensionality of $\tX$ and $\tA$, which scale badly with the number of regions and time points in our experiments.
We use the reverse Kullback-Leibler (r-KL) loss \citep{blei_review} to ensure this scalability.
Reverse Kullback-Leibler training aims at minimizing the KL divergence between the variational family $q$ and the unknown posterior distribution $p(\vq, \vr, \tH, \tX, \tA | \tY, \vc)$.
Since the posterior distribution is unknown, we cannot directly minimize this divergence.
Instead, we minimize and upper bound of that divergence, which amounts to maximizing the evidence lower bound (ELBO) under the variational distribution:
\begin{equation}
    \begin{aligned}
        \phi_\text{P}^*     &= \min_{\phi_\text{P}} \mathcal{L}_\text{P}^\text{r-KL} &  \\
                            &= \min_{\phi_\text{P}} \E_{\substack{\vq, \vr, \tH \sim q_\text{HP}\\\tX, \tA \sim q_\text{P}}}
                ( 
                  & \log p(\tY, \vc, \tX, \tA, \vq, \vr, \tH) \\
                  & \  & - \log q_\text{P}(\tX, \tA | \vq, \vr, \tH; \phi_\text{P})\\
                  & \  & - \log q_\text{HP}(\vq, \vr, \tH | \tY = \btt{Y}^\text{observed})
                &\ )            
    \end{aligned}
\end{equation}
where the estimator $q_\text{HP}$ ---described in the previous paragraph--- evaluated on the true observed signal $\btt{Y}^\text{observed}$ is used as the variational posterior for the hyper-parameters $\vq, \vr, \tH$.
$q_\text{HP}(\vq, \vr, \tH)$ is \textit{not} trained during this second phase to prevent mode collapse.\\ Theoretically, if the RF $\tH$ and the noise levels $\vq$ and $\vr$ were known, $\tX$ can be inferred via Wiener de-convolution. In turn, given the latent signals, $\tA$ can be inferred in closed-form via Bayesian linear regression.
Informed by those considerations, we choose a Gaussian variational family to approximate the exact $\tX$ and $\tA$ posterior distributions. To scale our method to hundreds of regions, we do not model the covariance between the different coefficients of $\tA$, hence the covariance matrix for the posterior of $\tA$ is modeled as diagonal. To obtain the mean and variance of the Gaussian approximations, we regress those from the value of the hyper-parameters $\tH, \vq, \vr$ using a simple MLP architecture.

\section{Synthetic experiments}
\label{sec:exp_synth}

The goal of this synthetic experiment is to validate our methodological claims.
h-VB avoids mode collapse ---the \textit{hurdle} described in Section  \ref{sec:inference}--- via the separate forward-KL training of the HP estimator, as described in Section \ref{sec:h-VB}.
In practice, this helps us recover the true uncertainty in the inference of the coupling matrix $\tA$.
We show that, in contrary, an off-the-shelf inference method underestimates the uncertainty in the coupling matrix $\tA$. \textbf{Data:} In this experiment, we use a synthetic sample from the MDSI generative model ---described in Section \ref{sec:methods}.
This means that the ground truth HRF $\tH$, variances levels $\vq, \vr$ and coupling $\tA$ are known.
We feed to two methods the the synthetic BOLD signal $\btt{Y}$. \textbf{Baseline:}
As a baseline for comparison, we use a variational Bayes method.
Contrary to h-VB, the entirety of the parameters ---including the hyper parameters $\tH, \vq, \vr$--- are inferred using the reverse-KL loss.
As a result, the baseline focuses on certain HRFs only and misses part of the solution space for $\tA$.
The baseline uses a Gaussian approximation for $\tA$ and $\tX$ (similar to h-VB).
The baselines approximates the posterior for $\vr$ and $\vq$ using log-Normal distributions.
The baseline approximates the posterior for $\alpha$ using a Normal distribution soft clipped to the range $]\-\pi/2;\pi/2[$ (using a rescaled sigmoid function).

\begin{figure}
    \centering
    \includegraphics[width=0.9\textwidth]{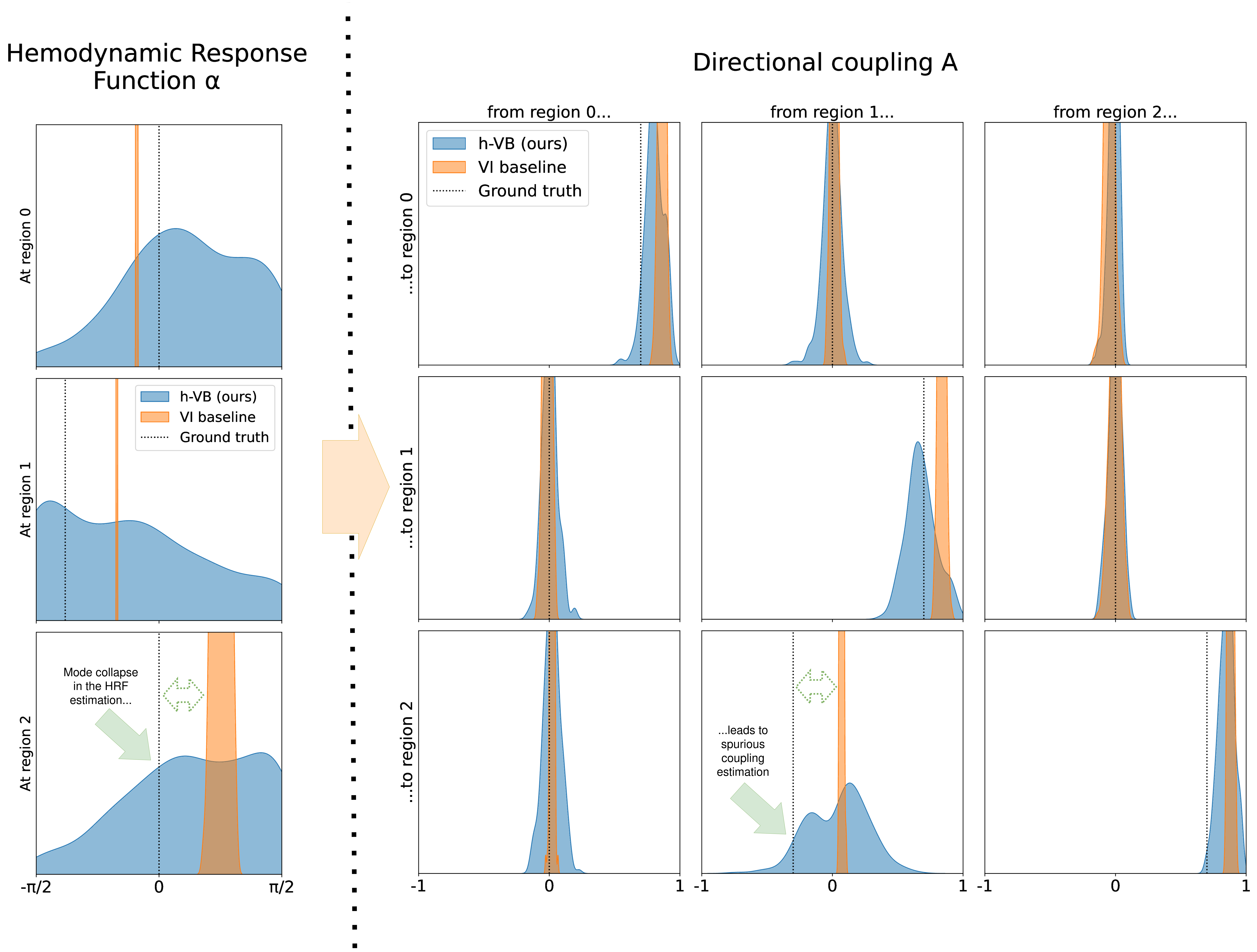}
    \caption{
        \textbf{Synthetic example inference}
        Posterior marginal distributions of the hyper-parameter $\alpha$ and the parameter $\tA$ ---as described in Section \ref{sec:h-VB}.
    }
    
    \label{fig:synthetic-HP}
    \label{fig:synthetic-A}
\end{figure}
\textbf{Hyper-Parameter inference: HRF and variance levels}
\ref{fig:synthetic-HP} (left) displays the $(\alpha, \vq, \vr)$ posterior distributions of h-VB and the baseline.
The baseline's posterior collapses to a small fraction of the posterior's support, thereby missing the ground truth parameters.
On the contrary, h-VB correctly recovers the entirety of the solution space.
Note that, without strong priors on the underlying signal $\tX$, inferring the HRF $\tH$ from the BOLD signal $\tY$ is ill-posed \citep{human_hrf}.
As a result, the support of h-VB's $\alpha$ posterior is very large. \textbf{Parameter inference: coupling matrix}
\ref{fig:synthetic-A} (right) displays the $\tA$ posterior distributions of h-VB and the baseline.
Since the baseline ignored most of the HRF $\tH$ solution space, it features peaked posteriors on spurious coupling values.
This means that the baseline outputs biased results with strong statistical confidence.
On the contrary, h-VB correctly considers all the different HRF scenarios that could have generated the BOLD signal $\tY$.
As an example, consider the only non-null coupling in this synthetic example: a strong negative coupling from region 1 to region 2.
Placing the threshold of the existence of a coupling at a $0.1$ value, the baseline outputs a 1\% chance of a positive coupling and a 0\% chance of a negative coupling (the ground truth).
On the contrary, h-VB outputs a 46\% chance for a positive coupling and a 30\% chance for a negative coupling (the ground truth).
h-VB helps the experimenter determine that, though a coupling is likely to exist between the 2 regions, inferring its sign is inconclusive.\\

In this experiment, we showed that off-the-shelf inference methods, though featuring a probabilistic output, can lead to over-estimated statistical confidence and spurious results.
h-VB, on the contrary, recovers the true uncertainty in the problem and can lead to more nuanced and richer conclusions.

\subsection{Mode collapse in practice: effect on ground truth coupling coverage}

\begin{table}[]
    \centering
    \begin{subfigure}[]{0.49\textwidth}
        \includegraphics[width=\textwidth]{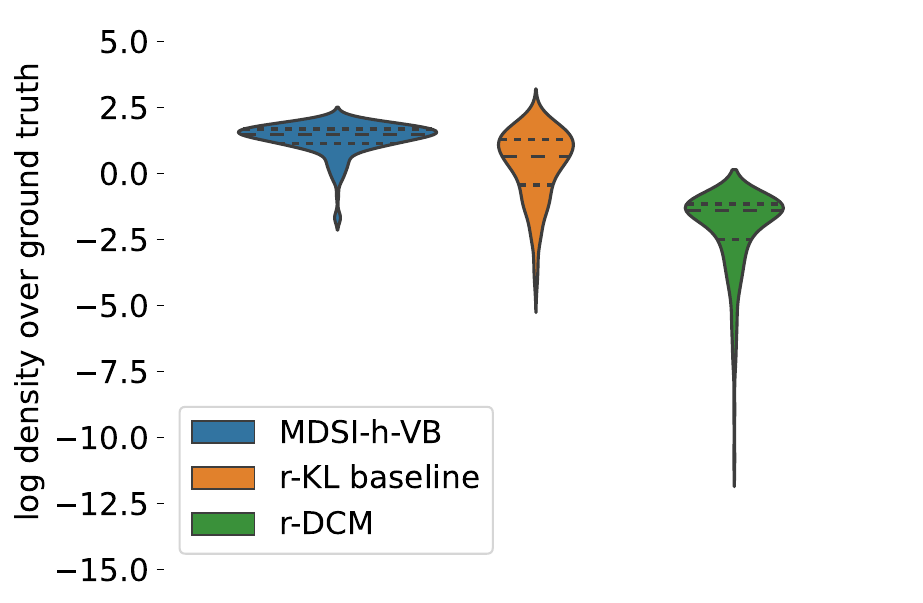}
    \end{subfigure}
    \hfill
    \begin{subfigure}[]{0.50\textwidth}
        \centering
        \begin{tabular}{|rr|ll|}
            Network     & \#nodes & \multicolumn{2}{c}{AUC}                              \\        
                        &                 & MDSI-h-VB                           & r-DCM               \\
            \hline
            1           & 5               & 0.82                           & \textbf{0.92}       \\
            2           & 5               & 0.79                           & \textbf{0.92}       \\
            3           & 5               & \textbf{0.95}                  & 0.88                \\
            4           & 10              & \textbf{0.94}                  & 0.83                \\
            5           & 5               & \textbf{0.91}                  & 0.70                \\
            6           & 8               & \textbf{0.93}                  & 0.88                \\
            7           & 6               & \textbf{0.82}                  & 0.72                \\
            8           & 8               & \textbf{0.89}                  & 0.78                \\
            9           & 9               & 0.82                           & \textbf{0.87}       \\
            Macaque & 28              & \textbf{0.92}                  & 0.76                \\
            Macaque & 91              & \textbf{0.90}                  & 0.89                \\
            \hline
            mean        &                 & \textbf{0.88} $\pm 0.05$     & 0.83$\pm 0.08$
        \end{tabular}
    \end{subfigure}
    \vspace{8pt}
    \caption{
        \textbf{Validation on synthetic data}
        \textit{\underline{Left:} MDS model, ground truth coupling posterior coverage}
        Posterior $\log$ density over the off-diagonal ground truth coupling coefficient values for different methods.
        \textit{\underline{Right:}} neurophysiological model. Methods are compared in terms of accuracy in connection detection. The macaque networks (bottom two lines) have a biologically relevant structure, obtained from tracer injection studies \citep{macaque_dataset}.
    }
    \label{tab:synth_macaque}
    \label{fig:mds_dataset}
\end{table}

This experiment validates statistically the effect of mode collapse as illustrated in \ref*{sec:exp_synth}. \textbf{Data:} We generate a synthetic dataset using the MDS model ---described in Section \ref*{sec:exp_synth}.
We generate 20 random networks with sparse coupling matrix.
Non-diagonal elements of $\tA$ have a 70\% chance to be null, 20\% to be $0.2$, and a 10\% chance to be $-0.2$.
For each network, we simulate 10 "subjects", corresponding to independent runs of the MDS model with the same coupling matrix. \textbf{Baseline:}
We use the same r-KL baseline as described in Section \ref*{sec:exp_synth}.
In addition, we compare to r-DCM, a recent scalable extension of DCM \citep{rdcm_1,rdcm_2}.
r-DCM uses a similar linear-coupling modeling as in the MDS model described in  Section \ref*{sec:methods}.
To invert its model, r-DCM uses Fourier analysis and Bayesian linear regression.
One major difference with MDSI-h-VB is that r-DCM does not take into account HRF variability, and assumes that every region is associated with the default HRF.
Mis-specification of the HRF is identified by \citet{rdcm_1} as one of their method's main limitations. \textbf{Metric:}
We leverage the probabilistic output of the compared methods.
Once the posterior is fitted, we compute the log density over the off-diagonal coupling coefficients.
This metric translates if the ground truth is statistically contained in the posterior distribution. \textbf{MDSI-h-VB recovers the ground truth coupling more reliably}
Results are visible in \ref{tab:synth_macaque} (left).
By taking into account HRF variability, yet avoiding mode collapse, MDSI-h-VB covers the full support of the posterior for the coupling matrix $\tA$.
This posterior thus contains the ground truth coupling value.
In contrast, the baselines feature more peaked posteriors that tend to "miss" the ground truth ---as illustrated in \ref*{fig:synthetic-A}.
The baseline's posterior density over the ground truth is thus lower than for MDSI-h-VB.

\section{Application on a neurophysiological synthetic dataset: connection detection}

The goal of this experiment is to validate our method on samples coming from a different generative model than the MDS.
The ground truth coupling is binary: either there is a positive coupling between regions, or there is no coupling (the strength of the coupling does not vary).
As a result, we test our method in terms of the accuracy of connection detection. \textbf{Data:}
We use synthetic data sampled using a neurophysiological process \citep{macaque_dataset}.
Underlying neural dynamics are simulated using the linear differential equation $\partial z / \partial t = \sigma \tA z + C u$, where $\tA$ denotes the ground-truth connectivity.
To simulate resting-state data, the $u$ input was modeled using a Poisson process for each of the regions.
The neuronal signals $z$ were then passed through the Balloon-Windkessel model \citep{friston_fmri} to obtain simulated BOLD data.
The networks 1-9 feature small-scale synthetic graphs, which vary widely in their density and number of cycles.
The Macaque networks consist of two larger graphs extracted from the macaque connectome. \textbf{Baseline:} We compare ourselves to a state-of-the-art directional coupling estimation method: r-DCM \citep{rdcm_1,rdcm_2}.
r-DCM is a Bayesian linear regression method in the Fourier domain, that does not take into account the HFR variability. r-DCM has been designed with scalability in mind, to be applied in the context of full-brain analysis. \textbf{Method:} For each method, network and subject, we infer the mean value of the coupling matrix $\tA$ posterior.
For each coefficient, we then compute a t-score across subjects.
We then feed that score to a binary logistic regression classifier.
We report the AUC of the classifier. \textbf{MDSI-h-VB connection detection accuracy is maintained as the number of nodes augments}
\ref{tab:synth_macaque} reports the connection detection AUC of MDSI-h-VB as the number of nodes in the network augments.
Both the Macaque cases feature several dozen nodes.
In addition, their ground truth connections are based on axonal connectivity derived from tracer injection studies \citep{macaque_dataset}.
As a result, the Macaque 91 setup is a good proxy for the performance of MDSI-h-VB on a full brain analysis as in \ref{sec:mdsi_scaling}.
In this challenging setup, MDSI-h-VB maintains an AUC of 0.90.

\section{Data-driven discovery of driving regions in human working memory}
\label{sec:mdsi_scaling}

\begin{figure}
    \centering
    \includegraphics[width=0.8\textwidth]{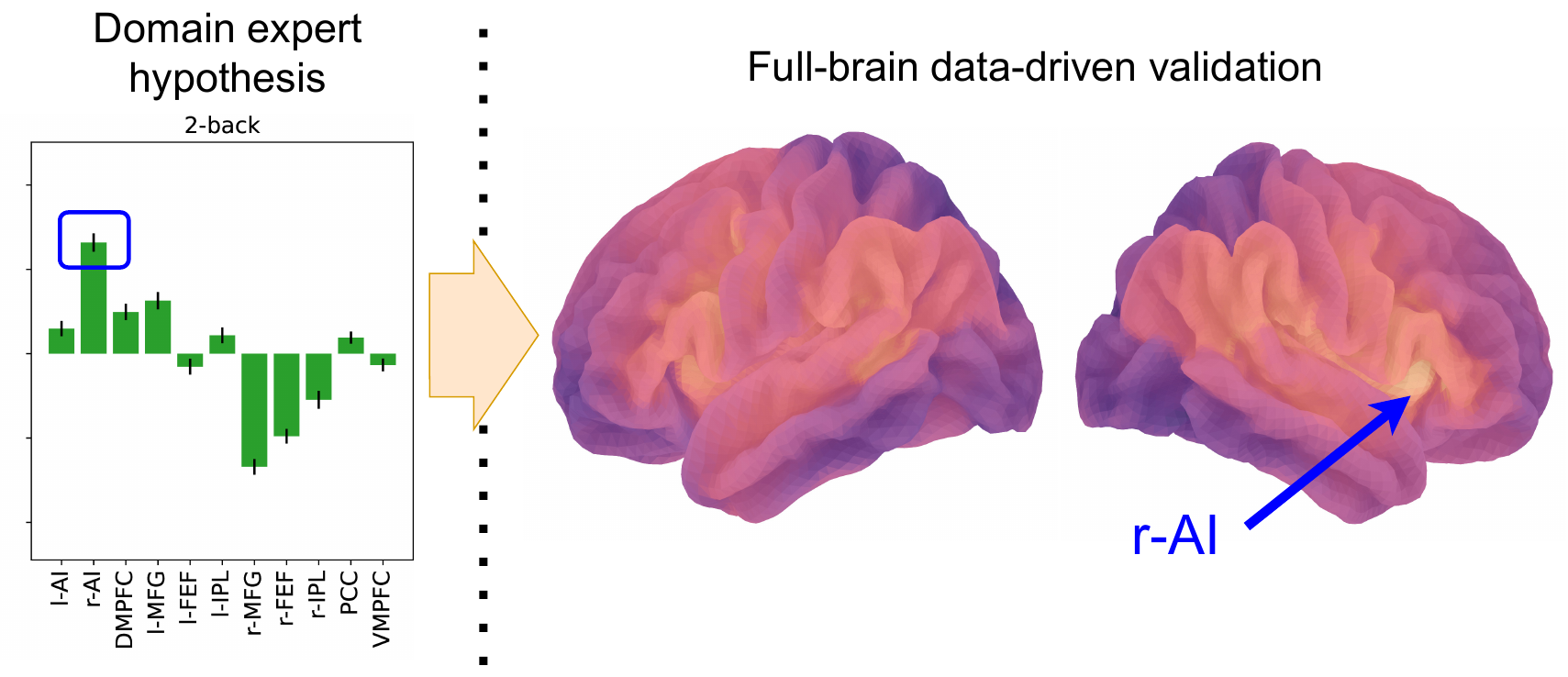}
    \caption{
        \textbf{Full-brain analysis confirms the driving role of the r-AI in working memory}
        \textit{On the left:} directed outflow analysis on 11 pre-selected ROIs.
        Working memory regions are selected by an expert, which can incur confounding from unobserved regions.
        \textit{On the right:} full-brain analysis, removing potential confounds.
        The r-AI was hypothesized to be a driving region in the 11-ROIs analysis (blue rectangle).
        The full-brain analysis confirms this analysis: the r-AI (blue arrow) appears as a hot spot of the directed outflow.
    }
    \label{fig:data_driven}
\end{figure}

This experiment leverage the inference speed and scalability of MDSI-h-VB to scale up our coupling analysis to the whole brain.
We apply MDSI-h-VB to data from the HCP dataset \citep{hcp}: 737 HCP subjects performing Working Memory tasks.
Subjects undergo three experimental conditions: a 2-back WM task, a 0-back WM task, and a baseline state between tasks.
For each subject, we use MDS-h-VB to infer the mean value of the coupling matrix $\tA$, one coupling matrix $\tA_c$ per condition $c$.
For each subject and region, we compute the directed outflow as the sum of the outwards coefficients minus the sum of the inwards coefficients:
\begin{equation}
    \begin{aligned}
        \forall c = 1..\conditions \ \forall m_1=1..\regions: \quad \text{directed\_outflow}_{c, m_1} = \sum_{\mathclap{m_2=1..M, m_2 \neq m_1}} a_{c,m_2,m_1} - \quad \sum_{\mathclap{m_2=1..M, m_2 \neq m_1}} a_{c,m_1,m_2}
    \end{aligned}
\end{equation}
The directed outflow translates whether a region mostly drives the signal of the rest of the brain (positive outflow) or if it is mostly driven by the rest of the brain (negative outflow).
We consider two choices of ROIs.
On the one hand, 11 pre-determined ROIs associated by experts to working memory \citep{mdsi_cai}.
On the other hand, 248 regions coming from a full brain parcellation \citep{brainnetome}.
Working on the whole brain yields a more data-driven approach than working with pre-determined ROIs.
We investigate the role of the r-AI as a driving hub in human working memory.
Results are visible in \ref{fig:data_driven}.
First, we reproduce previous findings \citep{mdsi_cai} in the 11-ROIs case, picking up the r-AI as the strongest outflow node.
Second, we validate those findings in a data-driven way, outlining the r-AI as a high outflow hub.
Working at the full brain scale, we ensure that our findings cannot be confounded by unobserved region ---as could be the case with pre-selected regions.

\section{Discussion}
In this paper, we introduced a method capable of reliably estimating large directed networks, on the order of hundred of nodes, from indirect and time-shifted measurement. In order to account from the ill-posedness of the inference, our Bayesian estimator automatically marginalizes-out the measurement parameters. This implies that our connectivity estimates are averaged over all plausible time-shifts and are therefore less sensitive to spurious inference due to time inversion. To perform this marginalization correctly in a variational framework, it is crucial to use a forward KL loss over these shift parameters as the conventional reversed KL loss is highly vulnerable to very severe underestimations of the uncertainty.

Our method open the door for reliable and scalable analysis of directed couplings in large networks, which could lead to breakthroughs in several fields. For example, reliable estimation of the directed causal structure of interconnection between brain regions from fMRI data can improve our understanding of functional brain networks both during rest and during cognitive activity \cite{bressler2010large, raichle2015brain, menon202320}. The approach can be potentially scaled to hundred of regions. However, the quadratic scaling in the number of possible connections poses a limit to the scalability to larger networks. Another limitation of the current approach is that it does not account for potential non-linearities in the dynamical and observation models. In the case of brain networks, the absence of threshold non-linearities complicates the interpretation of negative couplings, since in a linear models negative weights do not cause a suppression in activity. However, non-linear models can be straightforwardly integrated into our variational inference method and we leave this modification to future work.

\bibliography{references.bib}

\begin{thebibliography}{48}
\providecommand{\natexlab}[1]{#1}
\providecommand{\url}[1]{\texttt{#1}}
\expandafter\ifx\csname urlstyle\endcsname\relax
  \providecommand{\doi}[1]{doi: #1}\else
  \providecommand{\doi}{doi: \begingroup \urlstyle{rm}\Url}\fi

\bibitem[Ambrogioni et~al.(2019{\natexlab{a}})Ambrogioni, Ebel, Hinne, G{\"u}{\c{c}}l{\"u}, Gerven, and Maris]{ambrogioni2019spikecake}
L.~Ambrogioni, P.~Ebel, M.~Hinne, U.~G{\"u}{\c{c}}l{\"u}, M.~Gerven, and E.~Maris.
\newblock {SpikeCaKe}: Semi-analytic nonparametric bayesian inference for spike-spike neuronal connectivity.
\newblock In \emph{International Conference on Artificial Intelligence and Statistics}, 2019{\natexlab{a}}.

\bibitem[Ambrogioni et~al.(2019{\natexlab{b}})Ambrogioni, G{\"u}{\c{c}}l{\"u}, Berezutskaya, Borne, G{\"u}{\c{c}}l{\"u}t{\"u}rk, Hinne, Maris, and Gerven]{ambrogioni2019forward}
L.~Ambrogioni, U.~G{\"u}{\c{c}}l{\"u}, J.~Berezutskaya, E.~Borne, Y.~G{\"u}{\c{c}}l{\"u}t{\"u}rk, M.~Hinne, E.~Maris, and M.~Gerven.
\newblock Forward amortized inference for likelihood-free variational marginalization.
\newblock In \emph{International Conference on Artificial Intelligence and Statistics}, pages 777--786, 2019{\natexlab{b}}.

\bibitem[Andrieu et~al.(2003)Andrieu, de~Freitas, Doucet, and Jordan]{review_mcmc}
C.~Andrieu, N.~de~Freitas, A.~Doucet, and M.~I. Jordan.
\newblock An {Introduction} to {MCMC} for {Machine} {Learning}.
\newblock \emph{Machine Learning}, 50\penalty0 (1):\penalty0 5--43, Jan. 2003.

\bibitem[Barlow(1959)]{barlow1959autocorrelation}
J.~S. Barlow.
\newblock Autocorrelation and crosscorrelation analysis in electroencephalography.
\newblock \emph{IRE Transactions on Medical Electronics}, \penalty0 (3):\penalty0 179--183, 1959.

\bibitem[Blei et~al.(2017)Blei, Kucukelbir, and McAuliffe]{blei_review}
D.~M. Blei, A.~Kucukelbir, and J.~D. McAuliffe.
\newblock Variational {Inference}: {A} {Review} for {Statisticians}.
\newblock \emph{Journal of the American Statistical Association}, 112\penalty0 (518):\penalty0 859--877, Apr. 2017.

\bibitem[Bressler and Menon(2010)]{bressler2010large}
S.~L. Bressler and V.~Menon.
\newblock Large-scale brain networks in cognition: emerging methods and principles.
\newblock \emph{Trends in Cognitive Sciences}, 14\penalty0 (6):\penalty0 277--290, 2010.

\bibitem[Buckner(1998)]{buckner1998event}
R.~L. Buckner.
\newblock Event-related f{MRI} and the hemodynamic response.
\newblock \emph{Human Brain Mapping}, 6\penalty0 (5-6):\penalty0 373--377, 1998.

\bibitem[Cai et~al.(2021)Cai, Ryali, Pasumarthy, Talasila, and Menon]{mdsi_cai}
W.~Cai, S.~Ryali, R.~Pasumarthy, V.~Talasila, and V.~Menon.
\newblock Dynamic causal brain circuits during working memory and their functional controllability.
\newblock \emph{Nature Communications}, 12\penalty0 (1):\penalty0 3314, 2021.

\bibitem[Daunizeau et~al.(2012)Daunizeau, Stephan, and Friston]{daunizeau2012stochastic}
J.~Daunizeau, K.~E. Stephan, and K.~J. Friston.
\newblock Stochastic dynamic causal modelling of f{MRI} data: should we care about neural noise?
\newblock \emph{NeuroImage}, 62\penalty0 (1):\penalty0 464--481, 2012.

\bibitem[Fan et~al.(2016)Fan, Li, Zhuo, Zhang, Wang, Chen, Yang, Chu, Xie, Laird, et~al.]{brainnetome}
L.~Fan, H.~Li, J.~Zhuo, Y.~Zhang, J.~Wang, L.~Chen, Z.~Yang, C.~Chu, S.~Xie, A.~R. Laird, et~al.
\newblock The human brainnetome atlas: a new brain atlas based on connectional architecture.
\newblock \emph{Cerebral Cortex}, 26\penalty0 (8):\penalty0 3508--3526, 2016.

\bibitem[Fr{\"a}ssle et~al.(2017)Fr{\"a}ssle, Lomakina, Razi, Friston, Buhmann, and Stephan]{frassle2017regression}
S.~Fr{\"a}ssle, E.~I. Lomakina, A.~Razi, K.~J. Friston, J.~M. Buhmann, and K.~E. Stephan.
\newblock Regression dcm for f{MRI}.
\newblock \emph{NeuroImage}, 155:\penalty0 406--421, 2017.

\bibitem[Friston(2009)]{friston_fmri}
K.~J. Friston.
\newblock Modalities, modes, and models in functional neuroimaging.
\newblock \emph{Science}, 326\penalty0 (5951):\penalty0 399--403, 2009.

\bibitem[Friston et~al.(2003)Friston, Harrison, and Penny]{friston2003dynamic}
K.~J. Friston, L.~Harrison, and W.~Penny.
\newblock Dynamic causal modelling.
\newblock \emph{NeuroImage}, 19\penalty0 (4):\penalty0 1273--1302, 2003.

\bibitem[Frässle and Stephan(2022)]{rdcm_2}
S.~Frässle and K.~E. Stephan.
\newblock Test-retest reliability of regression dynamic causal modeling.
\newblock \emph{Network Neuroscience}, 6\penalty0 (1):\penalty0 135--160, Feb. 2022.

\bibitem[Frässle et~al.(2017)Frässle, Lomakina, Razi, Friston, Buhmann, and Stephan]{rdcm_1}
S.~Frässle, E.~I. Lomakina, A.~Razi, K.~J. Friston, J.~M. Buhmann, and K.~E. Stephan.
\newblock Regression {DCM} for {fMRI}.
\newblock \emph{NeuroImage}, 155, 2017.

\bibitem[Glover(1999)]{glover}
G.~H. Glover.
\newblock Deconvolution of impulse response in event-related {BOLD} f{MRI}.
\newblock \emph{NeuroImage}, 9\penalty0 (4):\penalty0 416--429, 1999.

\bibitem[Granger(1969)]{granger1969investigating}
C.~W.~J. Granger.
\newblock Investigating causal relations by econometric models and cross-spectral methods.
\newblock \emph{Econometrica}, pages 424--438, 1969.

\bibitem[Grienberger and Konnerth(2012)]{grienberger2012imaging}
C.~Grienberger and A.~Konnerth.
\newblock Imaging calcium in neurons.
\newblock \emph{Neuron}, 73\penalty0 (5):\penalty0 862--885, 2012.

\bibitem[Herculano-Houzel(2009)]{herculano2009human}
S.~Herculano-Houzel.
\newblock The human brain in numbers: a linearly scaled-up primate brain.
\newblock \emph{Frontiers in Human Meuroscience}, 3:\penalty0 857, 2009.

\bibitem[Kingma and Welling(2013)]{kingma2013auto}
D.~P. Kingma and M.~Welling.
\newblock Auto-encoding variational bayes.
\newblock \emph{arXiv preprint arXiv:1312.6114}, 2013.

\bibitem[Kravchenko-Balasha et~al.(2012)Kravchenko-Balasha, Levitzki, Goldstein, Rotter, Gross, Remacle, and Levine]{kravchenko2012fundamental}
N.~Kravchenko-Balasha, A.~Levitzki, A.~Goldstein, V.~Rotter, A.~Gross, F.~Remacle, and R.~D. Levine.
\newblock On a fundamental structure of gene networks in living cells.
\newblock \emph{Proceedings of the National Academy of Sciences}, 109\penalty0 (12):\penalty0 4702--4707, 2012.

\bibitem[Kucukelbir et~al.(2017)Kucukelbir, Tran, Ranganath, Gelman, and Blei]{kucukelbir2017automatic}
A.~Kucukelbir, D.~Tran, R.~Ranganath, A.~Gelman, and D.~M. Blei.
\newblock Automatic differentiation variational inference.
\newblock \emph{Journal of Machine Learning Research}, 18\penalty0 (14):\penalty0 1--45, 2017.

\bibitem[Laird et~al.(2008)Laird, Robbins, Li, Price, Cykowski, Narayana, Laird, Franklin, and F.]{laird2008modeling}
A.~R. Laird, J.~M. Robbins, K.~Li, L.~R. Price, M.~D. Cykowski, S.~Narayana, R.~W. Laird, C.~Franklin, and P.~T. F.
\newblock Modeling motor connectivity using tms/pet and structural equation modeling.
\newblock \emph{NeuroImage}, 41\penalty0 (2):\penalty0 424--436, 2008.

\bibitem[Lindquist et~al.(2009)Lindquist, Loh, Atlas, and Wager]{lindquist2009modeling}
M.~A. Lindquist, J.~M. Loh, L.~Y. Atlas, and T.~D. Wager.
\newblock Modeling the hemodynamic response function in f{MRI}: efficiency, bias and mis-modeling.
\newblock \emph{Neuroimage}, 45\penalty0 (1):\penalty0 S187--S198, 2009.

\bibitem[Loomba et~al.(2022)Loomba, Straehle, Gangadharan, Heike, Khalifa, Motta, Ju, Sievers, Gempt, Meyer, et~al.]{loomba2022connectomic}
S.~Loomba, J.~Straehle, V.~Gangadharan, N.~Heike, A.~Khalifa, A.~Motta, N.~Ju, M.~Sievers, J.~Gempt, H.~S. Meyer, et~al.
\newblock Connectomic comparison of mouse and human cortex.
\newblock \emph{Science}, 377\penalty0 (6602):\penalty0 eabo0924, 2022.

\bibitem[Menon(2023)]{menon202320}
V.~Menon.
\newblock 20 years of the default mode network: A review and synthesis.
\newblock \emph{Neuron}, 2023.

\bibitem[Mnih and Rezende(2016)]{mnih2016variational}
A.~Mnih and D.~Rezende.
\newblock Variational inference for monte carlo objectives.
\newblock In \emph{International Conference on Machine Learning}, pages 2188--2196. PMLR, 2016.

\bibitem[Papamakarios and Murray(2016)]{papamakarios2016fast}
G.~Papamakarios and I.~Murray.
\newblock Fast $\varepsilon$-free inference of simulation models with bayesian conditional density estimation.
\newblock \emph{Advances in Neural Information Processing Systems}, 29, 2016.

\bibitem[Papamakarios et~al.(2017)Papamakarios, Pavlakou, and Murray]{maf}
G.~Papamakarios, T.~Pavlakou, and I.~Murray.
\newblock Masked autoregressive flow for density estimation.
\newblock \emph{Advances in Neural Information Processing Systems}, 30, 2017.

\bibitem[Papamakarios et~al.(2019)Papamakarios, Nalisnick, Rezende, Mohamed, and Lakshminarayanan]{nf_review}
G.~Papamakarios, E.~Nalisnick, D.~J. Rezende, S.~Mohamed, and B.~Lakshminarayanan.
\newblock Normalizing {Flows} for {Probabilistic} {Modeling} and {Inference}.
\newblock \emph{arXiv:1912.02762}, Dec. 2019.

\bibitem[Raichle(2015)]{raichle2015brain}
M.~E. Raichle.
\newblock The brain's default mode network.
\newblock \emph{Annual review of neuroscience}, 38:\penalty0 433--447, 2015.

\bibitem[Ramsey et~al.(2010)Ramsey, Hanson, Hanson, Halchenko, Poldrack, and Glymour]{ramsey2010six}
J.~D. Ramsey, S.~J. Hanson, C.~Hanson, Y.~O. Halchenko, R.~A. Poldrack, and C.~Glymour.
\newblock Six problems for causal inference from f{MRI}.
\newblock \emph{NeuroImage}, 49\penalty0 (2):\penalty0 1545--1558, 2010.

\bibitem[Raut et~al.(2019)Raut, Mitra, Snyder, and Raichle]{raut2019time}
R.~V. Raut, A.~Mitra, A.~Z. Snyder, and M.~E. Raichle.
\newblock On time delay estimation and sampling error in resting-state f{MRI}.
\newblock \emph{NeuroImage}, 194:\penalty0 211--227, 2019.

\bibitem[Rouillard et~al.(2023)Rouillard, Le~Bris, Moreau, and Wassermann]{pavi}
L.~Rouillard, A.~Le~Bris, T.~Moreau, and D.~Wassermann.
\newblock {PAVI}: Plate-amortized variational inference.
\newblock \emph{Transactions on Machine Learning Research}, 2023.

\bibitem[Ryali et~al.(2011{\natexlab{a}})Ryali, Supekar, Chen, and Menon]{mdsi_method}
S.~Ryali, K.~Supekar, T.~Chen, and V.~Menon.
\newblock Multivariate dynamical systems models for estimating causal interactions in {fMRI}.
\newblock \emph{NeuroImage}, 54\penalty0 (2):\penalty0 807--823, Jan. 2011{\natexlab{a}}.

\bibitem[Ryali et~al.(2011{\natexlab{b}})Ryali, Supekar, Chen, and Menon]{ryali2011multivariate}
S.~Ryali, K.~Supekar, T.~Chen, and V.~Menon.
\newblock Multivariate dynamical systems models for estimating causal interactions in f{MRI}.
\newblock \emph{Neuroimage}, 54\penalty0 (2):\penalty0 807--823, 2011{\natexlab{b}}.

\bibitem[Ryali et~al.(2016{\natexlab{a}})Ryali, Chen, Supekar, Tu, Kochalka, Cai, and Menon]{ryali2016multivariate}
S.~Ryali, T.~Chen, K.~Supekar, T.~Tu, J.~Kochalka, W.~Cai, and V.~Menon.
\newblock Multivariate dynamical systems-based estimation of causal brain interactions in f{MRI}: Group-level validation using benchmark data, neurophysiological models and human connectome project data.
\newblock \emph{Journal of Neuroscience methods}, 268:\penalty0 142--153, 2016{\natexlab{a}}.

\bibitem[Ryali et~al.(2016{\natexlab{b}})Ryali, Shih, Chen, Kochalka, Albaugh, Fang, Supekar, Lee, and Menon]{ryali2016combining}
S.~Ryali, Y.-Y.~I. Shih, T.~Chen, J.~Kochalka, D.~Albaugh, Z.~Fang, K.~Supekar, H.~Lee, J., and V.~Menon.
\newblock Combining optogenetic stimulation and f{MRI} to validate a multivariate dynamical systems model for estimating causal brain interactions.
\newblock \emph{Neuroimage}, 132:\penalty0 398--405, 2016{\natexlab{b}}.

\bibitem[Sanchez-Romero et~al.(2018)Sanchez-Romero, Ramsey, Zhang, Glymour, Huang, and Glymour]{macaque_dataset}
R.~Sanchez-Romero, J.~Ramsey, K.~Zhang, M.~Glymour, B.~Huang, and C.~Glymour.
\newblock Estimating feedforward and feedback effective connections from f{MRI} time series: Assessments of statistical methods., 2018.

\bibitem[Shojaie and Fox(2022)]{shojaie2022granger}
A.~Shojaie and E.~B. Fox.
\newblock Granger causality: A review and recent advances.
\newblock \emph{Annual Review of Statistics and Its Application}, 9:\penalty0 289--319, 2022.

\bibitem[Steffener et~al.(2010)Steffener, Tabert, Reuben, and Stern]{hrf_circle}
J.~Steffener, M.~Tabert, A.~Reuben, and Y.~Stern.
\newblock Investigating hemodynamic response variability at the group level using basis functions.
\newblock \emph{Neuroimage}, 49\penalty0 (3):\penalty0 2113--2122, 2010.

\bibitem[Stephan and Roebroeck(2012)]{stephan2012short}
K.~E. Stephan and A.~Roebroeck.
\newblock A short history of causal modeling of f{MRI} data.
\newblock \emph{NeuroImage}, 62\penalty0 (2):\penalty0 856--863, 2012.

\bibitem[Taylor et~al.(2018)Taylor, Kim, and Ress]{human_hrf}
A.~J. Taylor, J.~H. Kim, and D.~Ress.
\newblock Characterization of the hemodynamic response function across the majority of human cerebral cortex.
\newblock \emph{NeuroImage}, 173:\penalty0 322--331, June 2018.

\bibitem[Van~Essen et~al.(2012)Van~Essen, Ugurbil, Auerbach, Barch, Behrens, Bucholz, Chang, Chen, Corbetta, Curtiss, Della~Penna, Feinberg, Glasser, Harel, Heath, Larson-Prior, Marcus, Michalareas, Moeller, Oostenveld, Petersen, Prior, Schlaggar, Smith, Snyder, Xu, and Yacoub]{hcp}
D.~C. Van~Essen, K.~Ugurbil, E.~Auerbach, D.~Barch, T.~E. Behrens, R.~Bucholz, A.~Chang, L.~Chen, M.~Corbetta, S.~W. Curtiss, S.~Della~Penna, D.~Feinberg, M.~F. Glasser, N.~Harel, A.~C. Heath, L.~Larson-Prior, D.~Marcus, G.~Michalareas, S.~Moeller, R.~Oostenveld, S.~E. Petersen, F.~Prior, B.~L. Schlaggar, S.~M. Smith, A.~Z. Snyder, J.~Xu, and E.~Yacoub.
\newblock {{T}he {H}uman {C}onnectome {P}roject: a data acquisition perspective}.
\newblock \emph{NeuroImage}, 62\penalty0 (4):\penalty0 2222--2231, Oct 2012.

\bibitem[Von~Bartheld et~al.(2016)Von~Bartheld, Bahney, and Herculano-Houzel]{von2016search}
C.~S. Von~Bartheld, J.~Bahney, and S.~Herculano-Houzel.
\newblock The search for true numbers of neurons and glial cells in the human brain: A review of 150 years of cell counting.
\newblock \emph{Journal of Comparative Neurology}, 524\penalty0 (18):\penalty0 3865--3895, 2016.

\bibitem[West et~al.(2020)West, Halliday, Bressler, Farmer, and Litvak]{west2020measuring}
T.~O. West, D.~M. Halliday, S.~L. Bressler, S.~F. Farmer, and V.~Litvak.
\newblock Measuring directed functional connectivity using non-parametric directionality analysis: Validation and comparison with non-parametric granger causality.
\newblock \emph{NeuroImage}, 218:\penalty0 116796, 2020.

\bibitem[White and Anderson(2005)]{white2005signaling}
M.~A. White and R.~G.~W. Anderson.
\newblock Signaling networks in living cells.
\newblock \emph{Annu. Rev. Pharmacol. Toxicol.}, 45:\penalty0 587--603, 2005.

\bibitem[Zivot and Wang(2006)]{zivot2006vector}
E.~Zivot and J.~Wang.
\newblock Vector autoregressive models for multivariate time series.
\newblock \emph{Modeling financial time series with S-PLUS{\textregistered}}, pages 385--429, 2006.

\end{thebibliography}

\end{document}